\documentclass{article}

\usepackage{arxiv}
\usepackage{float}
\usepackage[utf8]{inputenc} 
\usepackage[T1]{fontenc}    
\usepackage{hyperref}       
\usepackage{url}            
\usepackage{graphicx}
\usepackage{multirow}
\usepackage{booktabs}       
\usepackage{flafter}		
\usepackage{plantuml}
\graphicspath{ {./images/} }

\title{TableZa - A classical Computer Vision approach to Tabular Extraction}

\date{} 					
\author{ \href{https://orcid.org/0000-0003-1002-1828}{\includegraphics[scale=0.06]{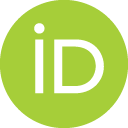}\hspace{1mm}Saumya Banthia}\\
	Synechron Innovation Lab (Bengaluru, India)\\
	\texttt{Saumya.Banthia@synechron.com}\\
	\And
	\href{https://orcid.org/0000-0002-9064-3362}{\includegraphics[scale=0.06]{orcid}\hspace{1mm}Anantha Sharma}\\
	Synechron Innovation Lab (Charlotte, NC)\\
	\texttt{Anantha.Sharma@synechron.com}\\
	\And
	{Ravi Mangipudi}\\
	Synechron Innovation Lab (San Francisco, CA)\\
	\texttt{ravi.vishwanath.m@gmail.com}\\
}

\renewcommand{\undertitle}{OpenCV Tabular Data Extraction}

\hypersetup{
pdftitle={OpenCV Tabular Data Extraction},
pdfsubject={cs.CV},
pdfauthor={Saumya Banthia},
pdfkeywords={computer vision, table, extraction, opencv},
}

\begin{document}
\maketitle

\begin{abstract}
	Computer aided Tabular Data Extraction has always been a very challenging and error prone task because it demands both Spectral and Spatial Sanity of data. In this paper we discuss an approach for Tabular Data Extraction in the realm of document comprehension. Given the different kinds of the Tabular formats that are often found across various documents, we discuss a novel approach using Computer Vision for extraction of tabular data from images or vector pdf(s) converted to image(s).
\end{abstract}


\section{Introduction}
\label{sec:intro}
Extracting tabular data from pdf documents is a growing field and there are many tools/packages like Camelot \cite{camelot}, Tabula \cite{tabula} to name a few which do a decent job extracting from text based (non-image) PDF documents. These tools fail if the table content is an image inside a pdf. This is mainly due to the fact that when an image is embedded inside the pdf, it does not contain any metadata of the character positions to read from thereby failing to accurately extract the content.

The techniques employed in extracting tabular data from images usually involves advanced machine techniques such as Deep learning \cite{cascadetabnet2020}, Graph Neural networks \cite{rethinkingGraphs} etc., and most of these solutions are computationally heavy and have a higher probability of failure while dealing with tightly constructed tables. One such open-source solution \cite{cascadetabnet2020} that was explored as part of the literature survey was fairly accurate and was able to detect tables, but missed out on detecting and extracting cells from the table. This gives us opportunity and scope to improve the detection and extraction by adding spectral and spatial components into our algorithm where our proposed method can be used. 

This paper focuses on the use of existing Computer Vision techniques and OCR tools to build two new approaches to solve the problem of image-based tabular data extraction. Our approach makes use of standardizing and dynamic adaptation which convert a wide variety of tables into a standard format, which can then be worked upon using a generalized approach making our method successful not only for the standard table but a large number of unique scenarios.

Although through the analysis of the results, it is easy to deduce that approach two is objectively better performing than approach one, but, techniques from both approaches can be combined to potentially produce even better results. This can be something that can be revisited in future.

\subsection{Scope}
The techniques highlighted in this document are not limited in scope of its applicability on image-based PDF's but can be used with images having tabular data in general. But whenever possible, text-based extraction of tabular data should be preferred over extraction from images (to reduce loss in accuracy that might result due to OCR or tabular structure estimation).

If the data happens to be exclusively available in image format, then the techniques highlighted in this paper become valuable.

\section{About the data}
\label{sec:data}
\begin{figure}[ht]
	\centering
	\includegraphics[scale=0.14]{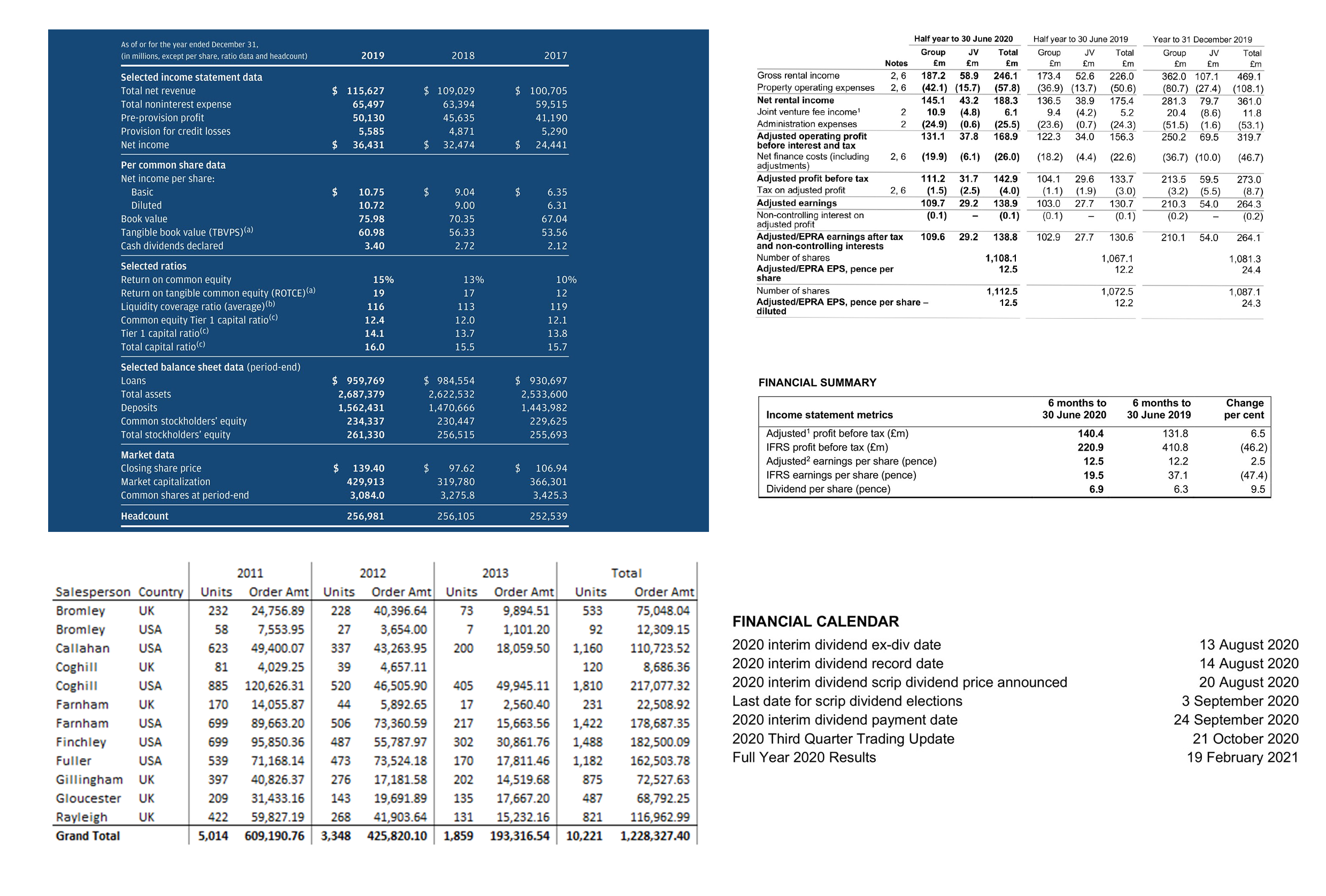}
	\caption{Examples of variety of tables occurring in the wild.}
	\label{fig:fig1}
\end{figure}

The tables we deal with are in various formats and structure [Figure \ref{fig:fig1}], a few variations have been listed below: 

\begin{itemize}
	\item Some have lines as column separators.
	\item Some have lines as row separators.
	\item Some have an arbitrary mix of the 2.
	\item Some have none (white-space acts as a separator).
	\item Apart from this there is vast variability in the structure of the tables themselves, which calls for a generalized approach.
	\item The color of the foreground and background can add an additional layer of challenge.
\end{itemize}

\subsection{Introduction to the tools}
The use of pre-existing tools has been done, so as not to re-invent the wheel. 

The following tools were used:
\begin{itemize}
	\item OpenCV \cite{2015opencv} (python-bindings) - OpenCV is a library of programming functions mainly aimed at real-time computer vision.
	\item Tesseract \cite{tesseract} (OCR) - Tesseract is an open source OCR or optical character recognition engine.
\end{itemize}

Similar techniques from other libraries can be used as a substitute for some of the libraries used in this case.
\raggedbottom
\pagebreak

\section{Methodology}
\label{sec:methodology}

\begin{figure}[ht]
	\centering
	\includegraphics[scale=0.65]{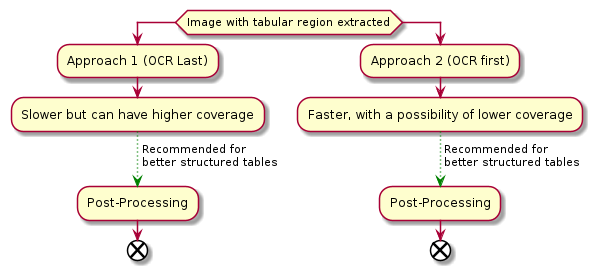}
	\caption{Overview of the 2 approaches}
	\label{fig:fig2}
\end{figure}

The main reasons why we ended up having two approaches are:
\begin{itemize}
	\item The accuracy of the 1st approach was found too low to be useful.
	\item The time being taken for a single pass for the 1st approach was quite high (about 10 seconds on a 16-core CPU).
\end{itemize}

We will now discuss the 2 approaches. Each has its own pros and cons.
Both approaches come with a presumption that the tabular area has already been detected (using one of the many available approaches to table detection) and the images being worked with are masked from or cropped into the tabular region of the image.

Post-processing is a recommended step for both techniques, which due to being common to both approaches, would be discussed towards the end.

\subsection{Approach One (OCR Last)}

\begin{figure}[ht]
	\centering
	\includegraphics[scale=0.4]{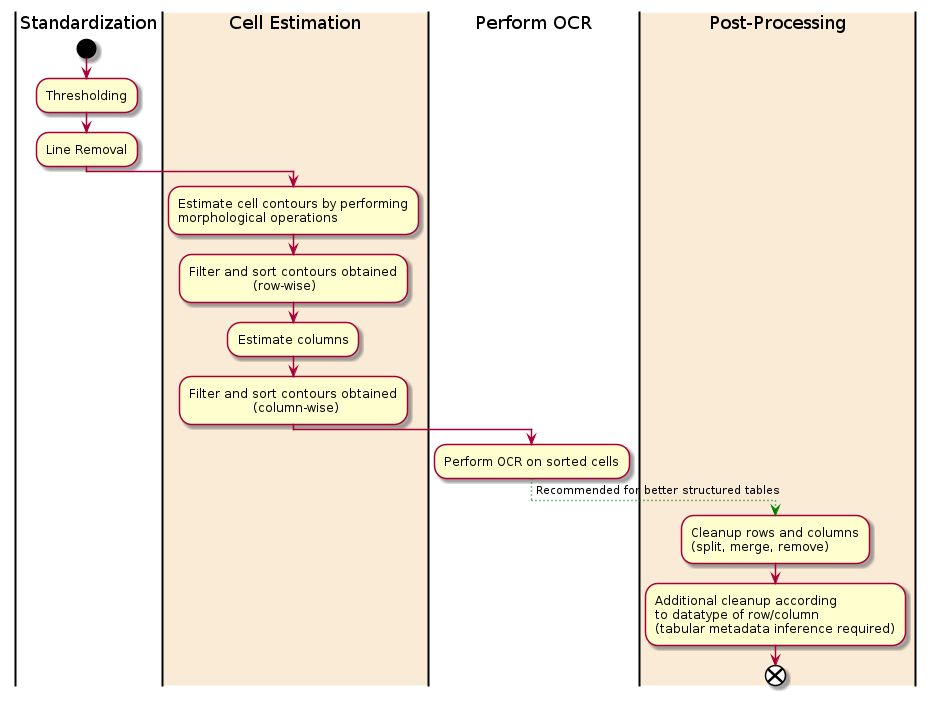}
	\caption{Approach One Flow}
	\label{fig:fig3}
\end{figure}

\subsubsection{Standardization}

\begin{figure}[ht]
	\centering
	\includegraphics[scale=0.75]{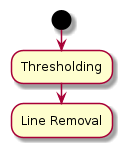}
	\caption{Standardization pipeline - Approach 1}
	\label{fig:fig4}
\end{figure}

Given the amount of variability in the formats of tables available, we would first need to standardize images, which makes applying further steps a uniform procedure for all table formats.

This includes 2 steps: 
\begin{itemize}
	\item Thresholding
	\item Line Removal
\end{itemize}

\begin{figure}[ht]
	\centering
	\includegraphics[scale=0.21]{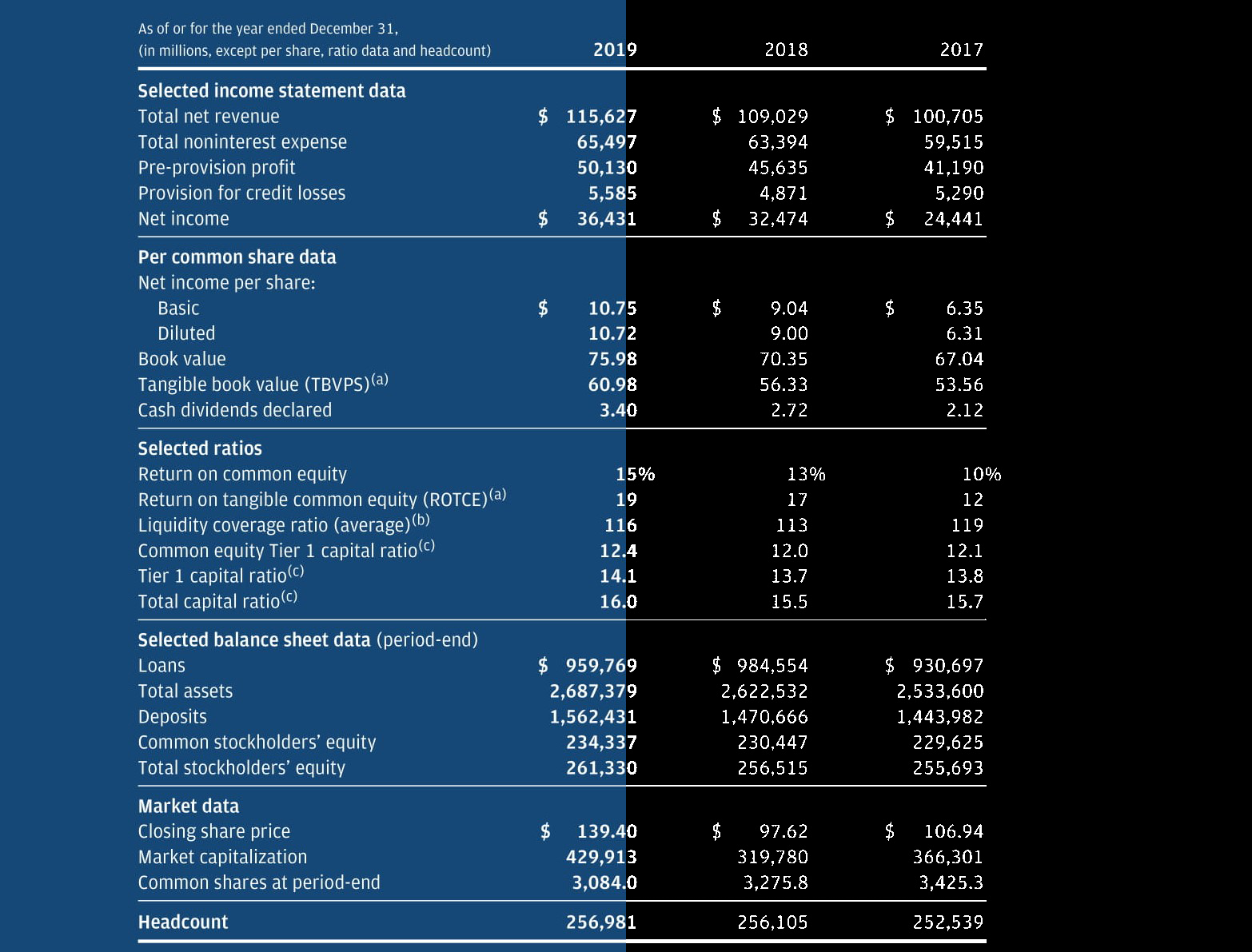}
	\caption{Thresholded image - Approach 1}
	\label{fig:fig5}
\end{figure}
\begin{figure}[ht]
	\centering
	\includegraphics[scale=0.19]{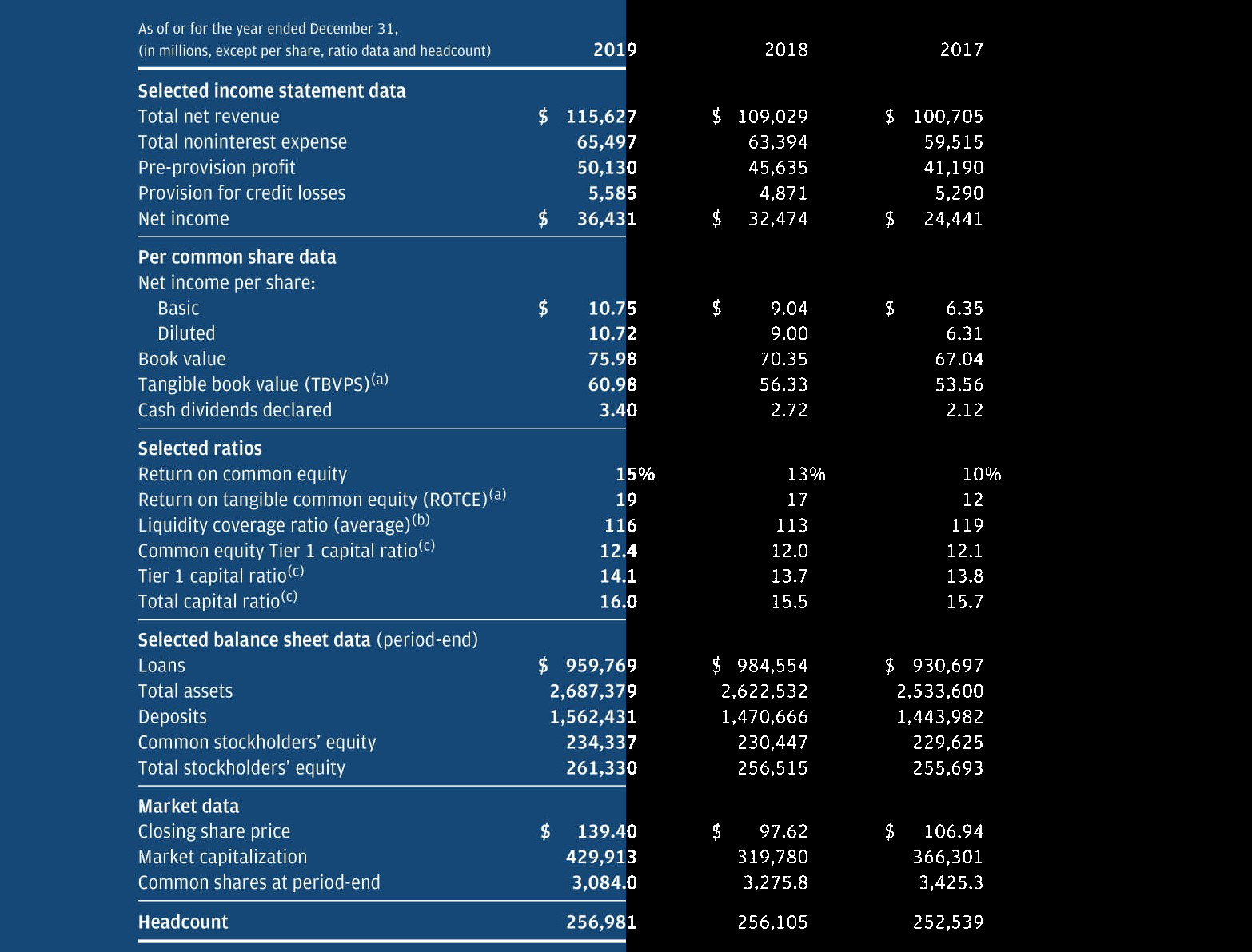}
	\caption{Lines removed - Approach 1}
	\label{fig:fig6}
\end{figure}

\paragraph{Thresholding}
Due to color variability, we might have the text (our foreground) sometimes be in white pixels and other times in black pixels when thresholded.

To standardize this, we first perform an initial thresholding to establish the foreground from the background. 

This can be achieved by doing a ratio check of the black and white pixels.
The reason why a ratio check would work for most documents is because, text (our foreground) however dense, has lower absolute pixel count compared to the background.

If the ratio is greater than 1 (black / white), it means the thresholding was performed correctly, else, we need to invert the thresholded image.
In doing so, we remove the color variability across tables.[Figure \ref{fig:fig5}]

\paragraph{Line Removal}

Line removal is another key step which involves morphological operations, namely contour formation and filtering. It is done both for removing horizontal and vertical lines. In all of the cases we examined, a structuring element of (40,1) - for horizontal lines and (1,40) - for vertical lines was found to be appropriate for this purpose. This can be made dynamic, so it changes with the size of the document, which is recommended. For all our other kernel size values, we use this technique to ensure wider coverage.

This would handle the variability of tables that have or do not have lines, and convert all tables into sets of rows and columns with only whitespace as separators.[Figure \ref{fig:fig6}]

\subsubsection{Cell Estimation}

\begin{figure}[ht]
	\centering
	\includegraphics[scale=0.75]{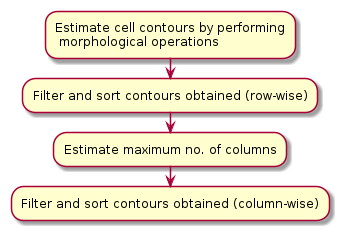}
	\caption{Cell Estimation pipeline - Approach 1}
	\label{fig:fig7}
\end{figure}

We dilated images containing tables both horizontally and vertically using a dynamically changing kernel size (using a formula which depends on the size of the document).
In our experimentation, we could not set a kernel size which could segment columns, without them bleeding into other columns (due to the vertical dilation), but a horizontal cell could be estimated through the same technique, without varying the filter size formula for various document size values.

One of the factors contributing to this error could be inconsistency of whitespace (when looking at data column-wise instead of row-wise).
It might also be due to the fact that the language of the documents we worked with were in English, which is written horizontally and therefore has relatively uniform horizontal spacing. 

\paragraph{Forming contours using Morphological Operations}

OpenCV offers a host of operations like dilation, opening and closing which can be performed to estimate and group cells at the row-level. The final result of this step is a list of contours and there approximate location (bounding-box coordinates). 

\begin{figure}[ht]
	\centering
	\includegraphics[scale=0.21]{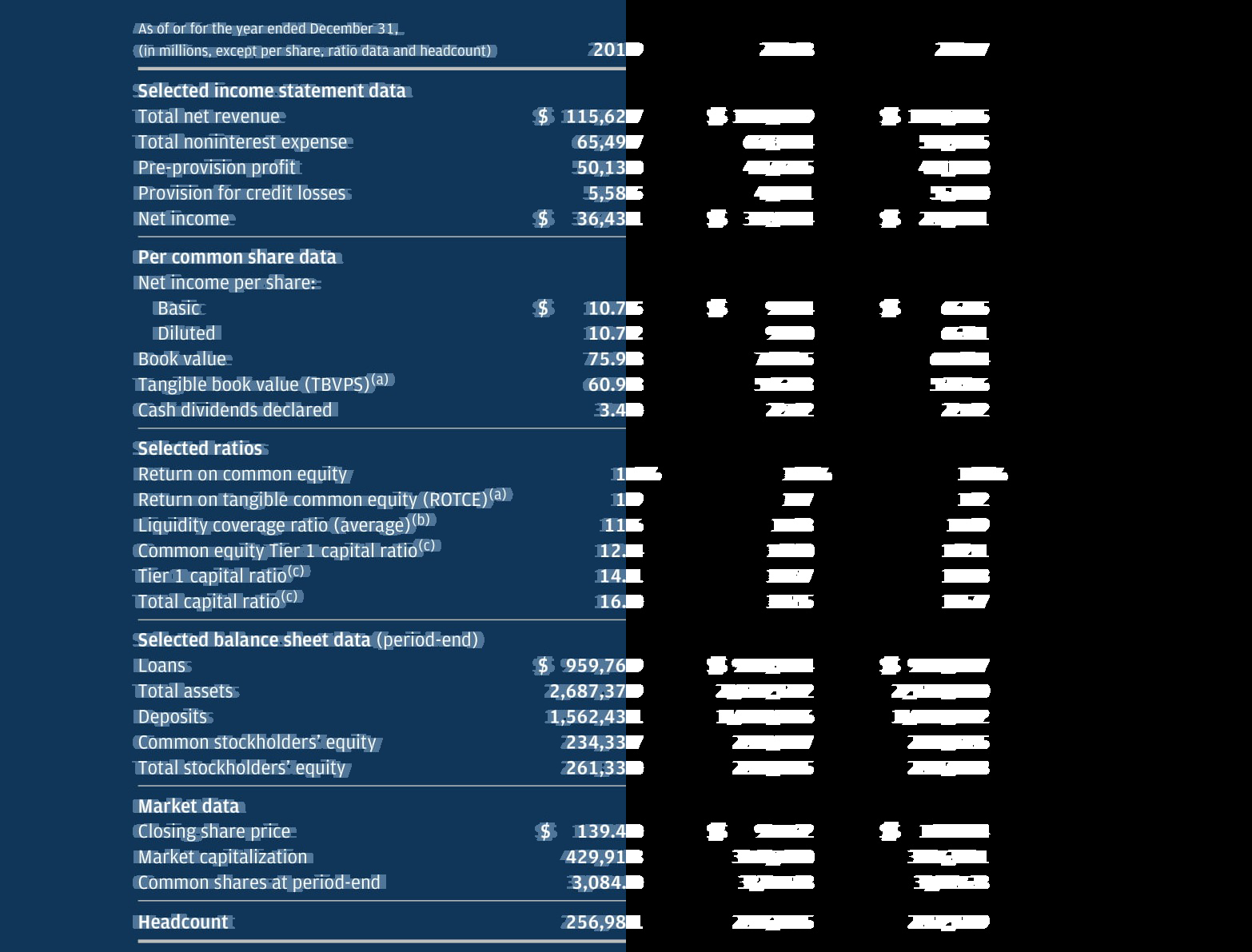}
	\caption{Contour Formation - Approach 1}
	\label{fig:fig8}
\end{figure}

\paragraph{Filter and sort contours (row-wise)}

This can be an optional step based on the criticality of the information being extracted.

This mostly handles smaller line segments which could not be removed while performing removal of lines as part of standardization.

The size of this threshold was made to change dynamically according to the size of the document to ensure coverage across different document sizes.

To do this, we took documents which contained these smaller elements (that were unnecessary) across different sizes and established a multiplier formula with respect to the document's size.

Filtered contours are then sorted from top-to-bottom, in this process assigning each contour to a certain row index. 

\paragraph{Estimating column bounds}

These row-wise sorted bounding boxes are then passed to the next step which sorts them from left-to-right (essentially sorting column-wise). We end up with a 2-d array. Using this array, we can now estimate number of columns, which can then be used to estimate column bounds.

Estimation of column bounds is done by taking the maximum width from a set of bounding boxes lying at the same column index (from the 2-d array).

\begin{figure}[ht]
	\centering
	\includegraphics[scale=0.21]{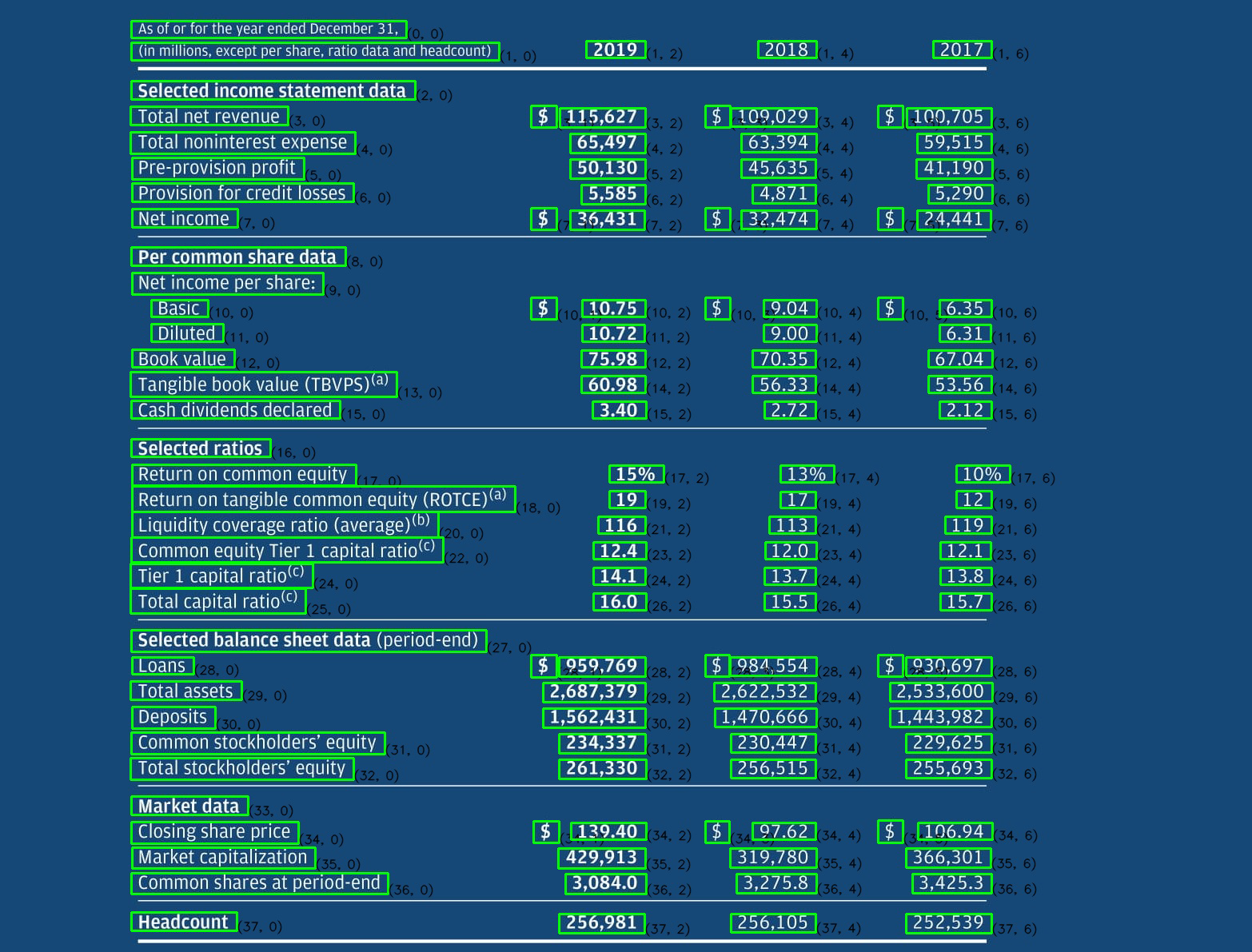}
	\caption{Cells demarcated with row and column index (zero-index) - Approach 1}
	\label{fig:fig9}
\end{figure}

There are 2 major challenges that arise here:
\begin{itemize}
	\item If elements in the same column are spread out, they get broken up into separate columns. (This is a relatively more common occurrence. Dollar signs (\$) in Figure \ref{fig:fig9} are an example of this problem.)
	\item If elements in different columns are close by, they might get merged into the same column. (In our experiments, this happened only once)
\end{itemize}

Both scenarios can be handled while post-processing, although splitting is a relatively challenging task.

\paragraph{Sorting contours (column-wise)}
\label{sssec: ref1}
Finally the estimated number of columns and the sorted list of bounding boxes is passed on for a final round of sorting, where each contour (and corresponding bounding box location) gets assigned a specific row and column index.

A simple algorithm goes through each element of every row, checks it against the set of column bounds estimated in the previous step and assigns the element to a specific row and column index. If 2 contours are assigned the same row and index values, they get merged into a single cell.

\subsubsection{Perform OCR}

The final step was to perform OCR on each of the estimated bounding box locations. In our testing, the results of this approach were not accurate enough to be considered useful (which is also showcased in Section \ref{sec:analysis}).

\raggedbottom
\pagebreak
\subsection{Approach 2 (OCR First)}

\begin{figure}[ht]
	\centering
	\includegraphics[scale=0.4]{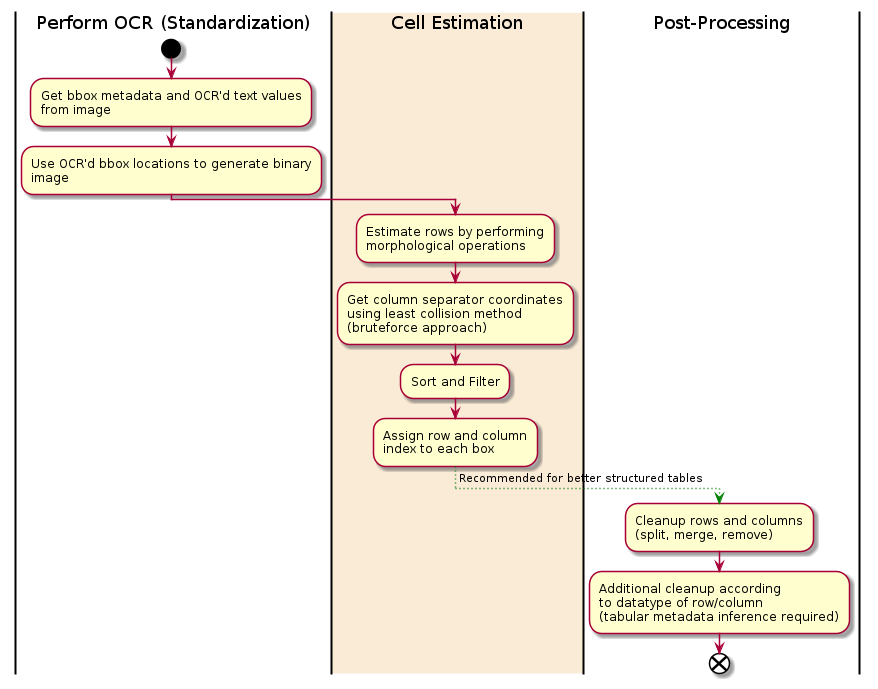}
	\caption{Approach Two Flow}
	\label{fig:fig10}
\end{figure}

\subsubsection{Perform OCR}

\begin{figure}[ht]
	\centering
	\includegraphics[scale=0.75]{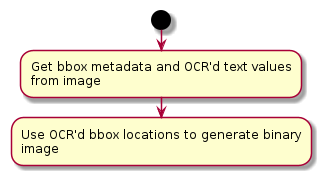}
	\caption{Perform OCR and Extract bounding box metadata - Approach 2}
	\label{fig:fig11}
\end{figure}

The first step in this approach happens to be the last step in the first approach and is also the reason for this approach being relatively much faster. A quick glance at swimlane diagrams of both approaches reveals that this technique does not require a standardization step, that is because we essentially offloaded that task to our OCR tool.

In essence, the bounding boxes extracted using this technique are only the ones which could be read by the OCR tool. An obvious con to this approach is missing out on text which the OCR engine could not ascertain to be text with a level of accuracy (which can be adjusted).

We use OCR tools (like Tesseract \cite{tesseract}) to get the bounding boxes as well as other metadata (confidence, row\_level, etc). The reason this approach is faster is because here the OCR happens on the image once for the whole image, which did not happen for the previous case. This negatively affects the previous approach in 2 ways:

\begin{itemize}
	\item Time Penalty - The previous approach was expensive (requiring us to run OCR for each cell), which may be mitigable through use of clever techniques.
	\item OCR Error Penalty - In the previous case, OCR is performed at the cell level, so if accuracy of OCR is bad (before/after processing image) then the potential rate of OCR error grows geometrically.
\end{itemize}

\subsubsection{Cell Estimation}

\begin{figure}[ht]
	\centering
	\includegraphics[scale=0.75]{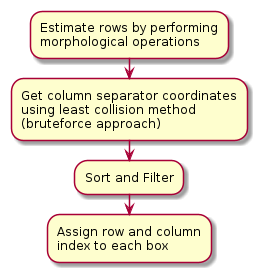}
	\caption{Cell Estimation pipeline - Approach 2}
	\label{fig:fig12}
\end{figure}

This is a common step between both approaches with functional differences which make up for the differences in approach.

\paragraph{Estimation of rows}
In this step, we perform morphological operations to dilate the bounding boxes we have received from the previous step. Here too, much like the previous approach we are using a formula which depends on the document size to dynamically resize the kernel used for dilating the detected bounding boxes. The result of this step is dilated bounding boxes.

\paragraph{Estimation of column bounds}
In this step, we use a technique we call the Least Collision method to brute-force column bounds for our table. A simplified version of the algorithm is summarized below:
\begin{itemize}
	\item Draw tightly spaced equi-distant vertical lines across the whole image.
	\item Keep lines having least collisions with bounding boxes (obtained in the previous step).
	\item Use a reducer function to iteratively remove really close-by lines.
	\item Lines remaining at the end can be assumed to be column bounds.
\end{itemize}

\paragraph{Sorting and Filtering}
This is same as the sorting and filtering performed for the previous approach. 

\paragraph{Assigning Row and Column index}
The sorted boxes can each now be assigned a particular row and column index of the table.

\raggedbottom
\pagebreak

\subsection{Post-Processing}

\begin{figure}[ht]
	\centering
	\includegraphics[scale=0.75]{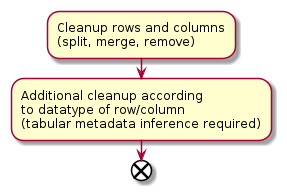}
	\caption{Post-Processing Pipeline - Both Approaches}
	\label{fig:fig13}
\end{figure}

Additional post-processing once the dataframe has been extracted is recommended to concatenate data which belongs together as expressed in section \ref{sssec: ref1} , along with some other specific issues that might arise as a result of using a generalized approach.

\subsubsection{Cleaning up rows and columns}
This step includes removal of empty lines, merging rows/columns which got split as a result of morphological operations (mostly dilation) and/or splitting of merged rows/columns (this is a trickier scenario to handle).

\subsubsection{Tabular Metadata based cleanup}
This involves contextual knowledge of the data being extracted, which can help estimate the datatype for certain columns or rows.

For example take the case of a column containing numeric data only, which has a numeric value \textbf{"4.577"} being wrongly OCR'd as \textbf{"4.5//"}. In this case if we have contextual knowledge that the column only contains numeric data, we can add a post-processing step to reduce errors like this.

\pagebreak
\raggedbottom

\section{Analysis}
\label{sec:analysis}
The two approaches outlined in the paper are compared against each other as well as a manually extracted ground truth. For the purpose of analysis, we have taken a small yet diverse set of 14 tables of different origin and types (bordered, border-less, colored, etc) and compared the differences in the generated dataframes across both models.

A con to the cleanup approach we applied was that it was universal in its application, instead of being applied only when necessary (which meant developing an algorithm to detect when it would be required). A classical approach would be counting rows and columns in the ground-truth (given that it is available) and using that as a basis. This smart cleanup technique was chosen to not be applied, assuming the real world scenario, where the actual number of rows and columns of the tables would not be prior knowledge.

Following are the metrics used for scoring:

\begin{itemize}
		\item Tabular structure:
		\begin{itemize}
			\item Row Count Delta
			\item Column Count Delta
		\end{itemize}
		\item Similarity Ratio between ground-truth dataframe and extracted dataframe - This uses longest string sequence matching for scoring
\end{itemize}

\begin{figure}[ht]
	\centering
	\includegraphics[scale=0.79]{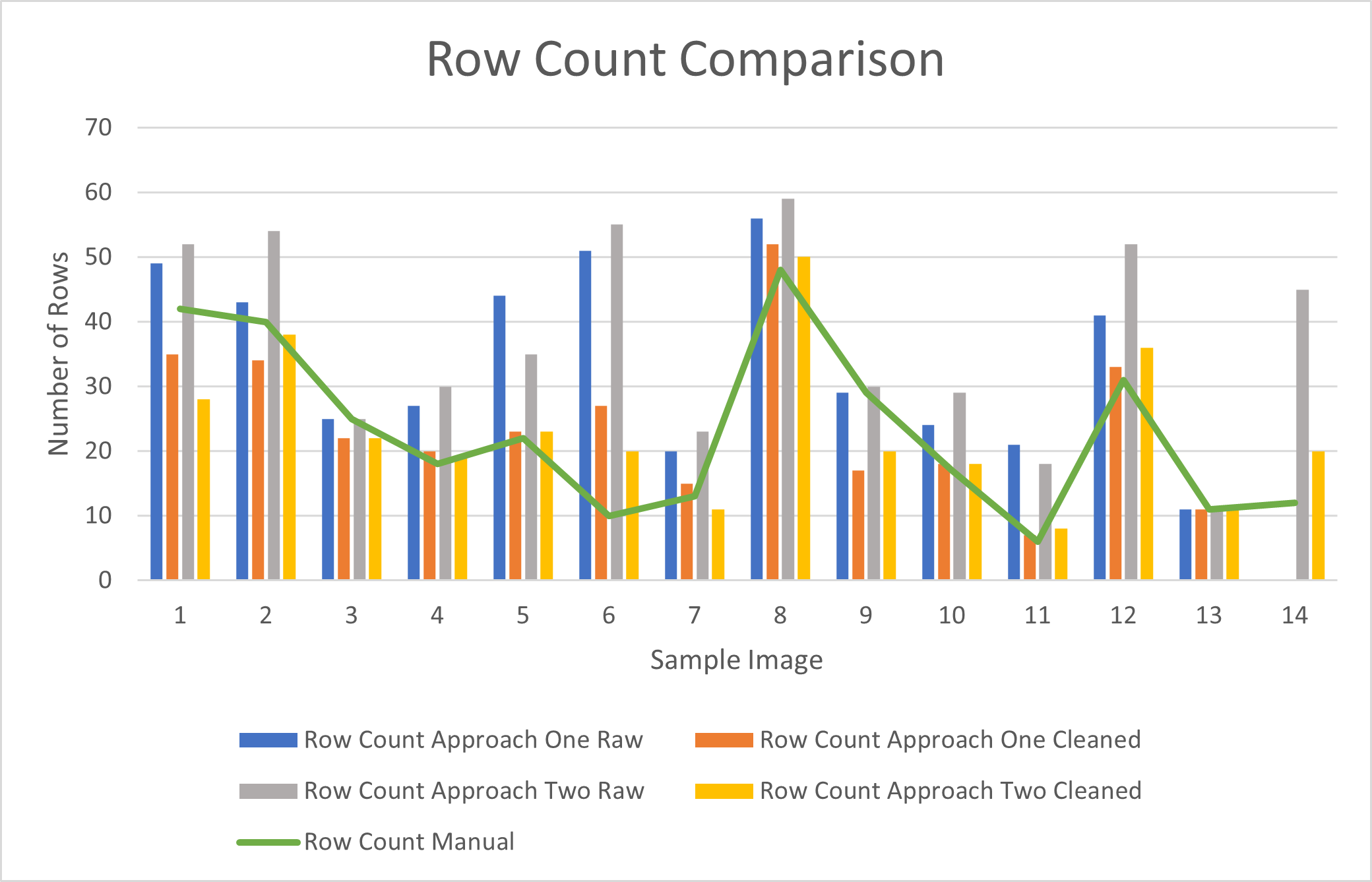}
	\caption{Row Count Comparison (Closer to the green line representing Row count Manual is better)}
	\label{fig:fig14}
\end{figure}

The row count gets better handled after cleanup of the data (which is a recommended step). To keep the cleanup simple, we only applied generalized cleanup approaches. So, improvements can be expected with better contextual understanding of the tables being extracted.

Observing the row count comparison chart [Figure \ref{fig:fig14}], you would notice for the majority of the cases, the cleaned versions of the extracted tables are closer in row count to the manually extracted version, when compared to the raw versions. The maximum overall error for raw and cleaned versions was 45, while average error was 9 (rounded up).

Column count comparison is similar to row count comparison. Although the data produced using this as a metric, was not quite useful relative to row count. This was because in majority of the cases the column count was accurate, with maximum overall error being 2, while average error being 1 (rounded up).

Another metric called Similarity Ratio was used. The reason we chose this instead of performing cell-wise scoring is because cell-wise mapping was a tricky thing to achieve in this case, given the vast difference in number of rows and columns in the tables that were produced by the two approaches. That would involve estimation (which may or may not be correct) of row(s)-column(s) combinations for cell-wise mapping, which could further lead to erroneous analysis.

Following observations were made upon analysis of the ground-truth tables compared against the ones generated using the 2 approaches:

\begin{itemize}
	\item Average row delta (values rounded up to nearest whole number) compared to the manual versions are:
	\begin{itemize}
		\item Approach One: 5 (went from 10 (raw) to 5 (cleaned))
		\item Approach Two: 9 (went from 14 (raw) to 5 (cleaned))
	\end{itemize}
	\item Following significant changes in similarity ratio were observed:
	\begin{itemize}
		\item Highest drop in similarity ratio (from raw to cleaned) were reported as follows:
		\begin{itemize}
			\item Approach One: -0.194860359 (went from 0.442922374 (raw) to 0.248062016 (cleaned))
			\linebreak
			(observation 9 in the table)
			\item Approach Two: -0.118603012 (went from 0.99054665 (raw) to 0.871943639 (cleaned))
			\linebreak
			(observation 9 in the table)
		\end{itemize}
		In the above cases (on the same observation), cleanup resulted in a reduced accuracy which means a developing a cleanup heuristic could help improve results.
		\item Highest increase in similarity ratio (from raw to cleaned) were reported as follows:
		\begin{itemize}
			\item Approach One: 0.094235522 (went from 0.276331131 (raw) to 0.370566653	(cleaned))\linebreak
			(observation 6 in the table)
			\item Approach Two: 0.533923958	(went from 0.10547504 (raw) to 0.639398998 (cleaned))
			\linebreak
			(observation 7 in the table)
		\end{itemize}
		In the above cases, the row count difference compared to the manual version have a high delta as well, as evident from [Figure \ref{fig:fig14}] and [Table \ref{tab:table1}].
	\end{itemize}
\end{itemize}

\begin{table}[htp]
	\centering
	\resizebox{\textwidth}{!}{%
	\begin{tabular}{@{}|l|l|l|l|l|l|l|l|l|l|l|l|l|l|l|@{}}
	\toprule
	\multicolumn{1}{|c|}{\multirow{3}{*}{Name}} &
	  \multicolumn{5}{c|}{Row   Count} &
	  \multicolumn{5}{c|}{Column   Count} &
	  \multicolumn{4}{c|}{Similarity Ratio} \\ \cmidrule(l){2-15} 
	\multicolumn{1}{|c|}{} &
	  \multicolumn{1}{c|}{\multirow{2}{*}{Manual}} &
	  \multicolumn{2}{c|}{Approach One} &
	  \multicolumn{2}{c|}{Approach Two} &
	  \multicolumn{1}{c|}{\multirow{2}{*}{Manual}} &
	  \multicolumn{2}{c|}{Approach One} &
	  \multicolumn{2}{c|}{Approach Two} &
	  \multicolumn{2}{c|}{Approach One} &
	  \multicolumn{2}{c|}{Approach Two} \\ \cmidrule(lr){3-6} \cmidrule(l){8-15} 
	\multicolumn{1}{|c|}{} &
	  \multicolumn{1}{c|}{} &
	  \multicolumn{1}{c|}{Raw} &
	  \multicolumn{1}{c|}{Cleaned} &
	  \multicolumn{1}{c|}{Raw} &
	  \multicolumn{1}{c|}{Cleaned} &
	  \multicolumn{1}{c|}{} &
	  \multicolumn{1}{c|}{Raw} &
	  \multicolumn{1}{c|}{Cleaned} &
	  \multicolumn{1}{c|}{Raw} &
	  \multicolumn{1}{c|}{Cleaned} &
	  \multicolumn{1}{c|}{Raw} &
	  \multicolumn{1}{c|}{Cleaned} &
	  \multicolumn{1}{c|}{Raw} &
	  \multicolumn{1}{c|}{Cleaned} \\ \midrule
	1  & 42 & 49 & 35 & 52 & 28 & 4  & 5  & 5  & 6  & 6  & 0.458498 & 0.446414 & 0.371196 & 0.548468 \\ \midrule
	2  & 40 & 43 & 34 & 54 & 38 & 7  & 7  & 7  & 7  & 7  & 0.288063 & 0.260557 & 0.471303 & 0.408098 \\ \midrule
	3  & 25 & 25 & 22 & 25 & 22 & 6  & 7  & 7  & 6  & 6  & 0.373966 & 0.366711 & 0.896988 & 0.827837 \\ \midrule
	4  & 18 & 27 & 20 & 30 & 19 & 6  & 8  & 8  & 6  & 6  & 0.621712 & 0.629765 & 0.860186 & 0.982232 \\ \midrule
	5  & 22 & 44 & 23 & 35 & 23 & 3  & 4  & 4  & 3  & 3  & 0.404179 & 0.440816 & 0.871709 & 0.819755 \\ \midrule
	6  & 10 & 51 & 27 & 55 & 20 & 8  & 9  & 9  & 8  & 8  & 0.276331 & 0.370567 & 0.334391 & 0.631958 \\ \midrule
	7  & 13 & 20 & 15 & 23 & 11 & 12 & 14 & 14 & 12 & 12 & 0.310725 & 0.20562  & 0.105475 & 0.639399 \\ \midrule
	8  & 48 & 56 & 52 & 59 & 50 & 8  & 10 & 10 & 8  & 7  & 0.198873 & 0.203398 & 0.755486 & 0.799557 \\ \midrule
	9  & 29 & 29 & 17 & 30 & 20 & 3  & 3  & 3  & 3  & 3  & 0.442922 & 0.248062 & 0.990547 & 0.871944 \\ \midrule
	10 & 17 & 24 & 18 & 29 & 18 & 11 & 11 & 11 & 10 & 10 & 0.36662  & 0.310531 & 0.540923 & 0.554007 \\ \midrule
	11 & 6  & 21 & 7  & 18 & 8  & 3  & 3  & 3  & 3  & 3  & 0.796532 & 0.796671 & 0.822278 & 0.870558 \\ \midrule
	12 & 31 & 41 & 33 & 52 & 36 & 8  & 8  & 8  & 8  & 8  & 0.643305 & 0.687225 & 0.540335 & 0.508207 \\ \midrule
	13 & 11 & 11 & 11 & 11 & 11 & 3  & 3  & 3  & 3  & 3  & 0.668513 & 0.668513 & 0.998935 & 0.998935 \\ \midrule
	14 & 12 & NA & NA & 45 & 20 & 9  & NA & NA & 8  & 8  & NA       & NA       & 0.548113 & 0.696745 \\ \bottomrule
	\end{tabular}%
	}
	\linebreak
	\caption{Comparison Table (Note: NA represents data not available)}
\label{tab:table1}
\end{table}

\begin{figure}[ht]
	\centering
	\includegraphics[scale=0.79]{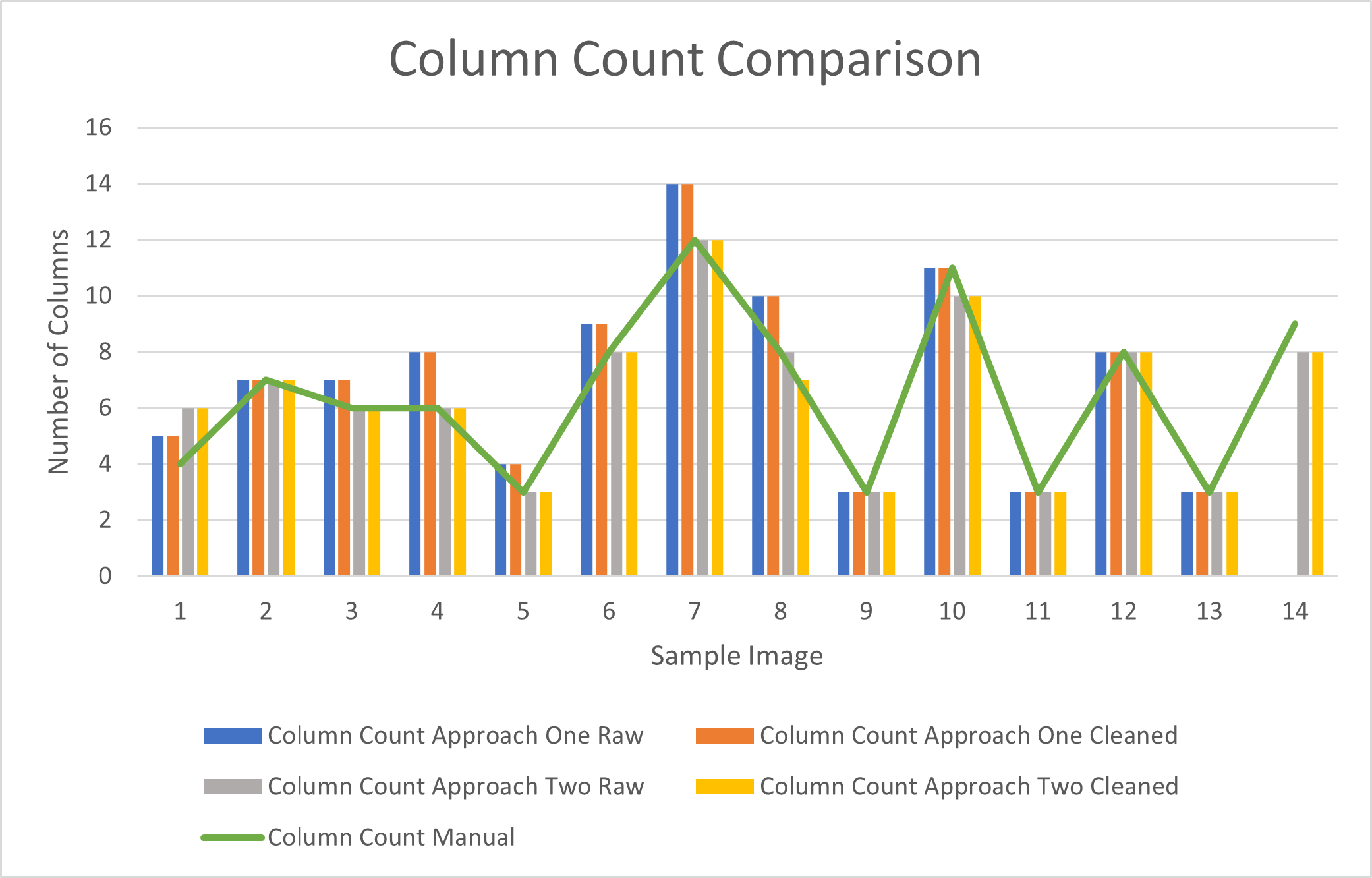}
	\caption{Column Count Comparison (Closer to the green line representing Column count Manual is better)}
	\label{fig:fig15}
\end{figure}

\begin{figure}[ht]
	\centering
	\includegraphics[scale=0.79]{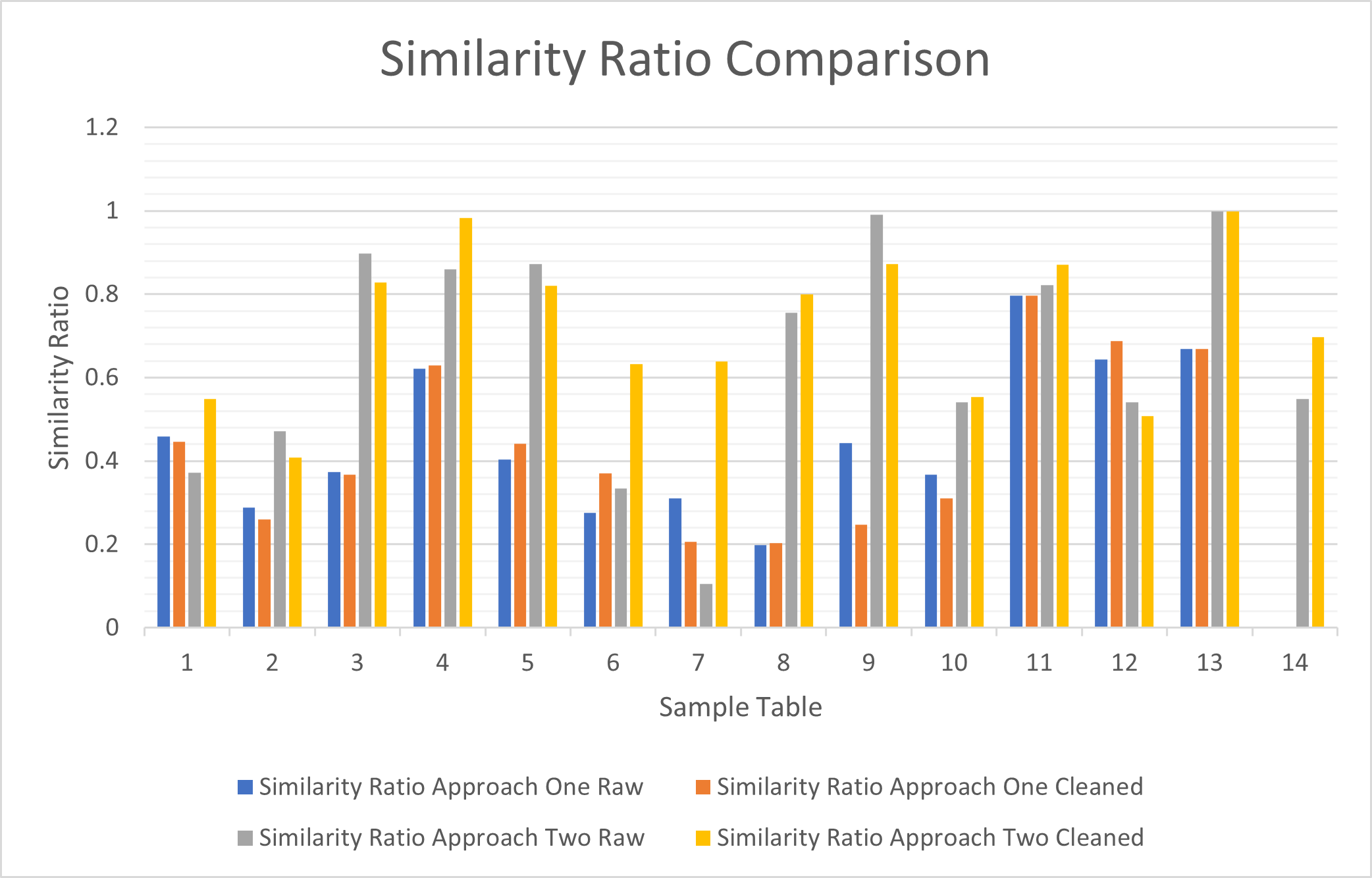}
	\caption{Similarity Ratio Comparison - Higher is better (Closer to 1)}
	\label{fig:fig16}
\end{figure}

\pagebreak
\raggedbottom

With the similarity ratio, a distinction gets drawn between approach one and two. The second approach (OCR First) seems to perform better than both raw and cleaned versions of the first approach. Therefore, apart from being significantly faster, the second approach also leads to better results overall.

On average, the best results from the first approach had a similarity ratio of 46.4\%. This gets easily taken over by even the average of the raw version of the second approach, which stood at 65\%. Whereas the best results from the second approach take the first place with an average similarity ratio of 74.9\%. This can be increased to 75.9\%, if we choose the best out of both the approaches. A smart cleanup algorithm (not a part of this paper) can help achieve this.

\pagebreak
\raggedbottom

\section{Conclusion}
\label{sec:conclusion}

The approaches highlighted in this document are two of many possible ways to extract tables from an image.

One of the ways in which this approach can be seen as unique is in its dependence solely on Computer Vision techniques for extraction of tables from images.

The obvious pros of this approach are: 

\begin{itemize}
	\item The results of the approach are completely interpretable.
	\item The algorithm is general enough to apply to a large variety of tabular images.
\end{itemize}

And the obvious cons are:

\begin{itemize}
	\item It is not (in its current form) a learning based approach, so the results are the best they can ever be.
	\item A generalized approach can be too vague at times, thus, requiring large amounts of pre/post-processing.
\end{itemize}

\section{Future Work}
\label{sec:futurework}
\begin{itemize}
	\item Contrary to one of the cons stated in the previous section, this generalized approach can be tuned to work better with certain scenarios.
	\item Training a custom OCR model can lead to improvement in OCR accuracy.
	\item Using combination of image transformation techniques can result in improved accuracy on OCR.
	\item Developing a cleanup heuristic for extracted tables, can help improve results.
	\item Contextual understanding of tabular data (leveraging metadata or a-priori knowledge about the data being extracted) can help improve the post-processing pipeline.
\end{itemize}

\bibliographystyle{unsrt}
\bibliography{references}

\end{document}